\documentclass[fleqn,10pt]{wlscirep}
\usepackage[utf8]{inputenc}
\usepackage[T1]{fontenc}
\usepackage{multirow}

\title{Quantized Large Language Models in Biomedical Natural Language Processing: Evaluation and Recommendation}

\author[1,2]{Zaifu Zhan}
\author[2]{Shuang Zhou}
\author[2]{Min Zeng}
\author[2]{Kai Yu}
\author[3]{Meijia Song}
\author[2]{Xiaoyi Chen}
\author[2]{Jun Wang}
\author[2]{Yu Hou}
\author[2,*]{Rui Zhang}
\affil[1]{Department of Electrical and Computer Engineering, University of Minnesota, Minneapolis, MN, USA}
\affil[2]{Division of Computational Health Sciences, Department of Surgery, University of Minnesota, Minneapolis, MN, USA}
\affil[3]{School of Nursing, University of Minnesota, Minneapolis, MN, USA}
\affil[*]{Corresponding author: Rui Zhang (\href{zhan1386@umn.edu}{zhan1386@umn.edu})}

\begin{abstract}
Large language models have demonstrated remarkable capabilities in biomedical natural language processing, yet their rapid growth in size and computational requirements present a major barrier to adoption in healthcare settings where data privacy precludes cloud deployment and resources are limited. In this study, we systematically evaluated the impact of quantization on 12 state-of-the-art large language models, including both general-purpose and biomedical-specific models, across eight benchmark datasets covering four key tasks: named entity recognition, relation extraction, multi-label classification, and question answering. We show that quantization substantially reduces GPU memory requirements—by up to 75\%—while preserving model performance across diverse tasks, enabling the deployment of 70B-parameter models on 40GB consumer-grade GPUs. In addition, domain-specific knowledge and responsiveness to advanced prompting methods are largely maintained. These findings provide significant practical and guiding value, highlighting quantization as a practical and effective strategy for enabling the secure, local deployment of large yet high-capacity language models in biomedical contexts, bridging the gap between technical advances in AI and real-world clinical translation.
\end{abstract}

\begin{document}

\flushbottom
\maketitle

%
%
\thispagestyle{empty}

\section{Introduction}

In recent years, large language models (LLMs) have rapidly permeated various aspects of biomedical natural language processing (NLP)~\cite{zhou2025large,chen2025benchmarking,zhan2025evaluation, zhou2025automating}, demonstrating remarkable capabilities across a wide range of tasks. The emergence of increasingly powerful general-purpose LLMs~\cite{achiam2023gpt,team2023gemini,liu2024deepseek}, along with a growing number of domain-adapted models specifically fine-tuned for biomedical applications~\cite{luo2022biogpt,tu2024towards,sellergren2025medgemma,zhan2025ramie}, has significantly advanced the field. Through techniques such as prompt engineering~\cite{brown2020language,zaghir2024prompt,wang2025data} and retrieval-augmented generation (RAG)~\cite{lewis2020retrieval,zhan2025retrieval,zhan2025mmrag}, LLMs are now able to generalize effectively even to unseen biomedical scenarios without requiring additional training.
While these developments are promising, they come at the cost of ever-growing model sizes and computational demands~\cite{kaplan2020scaling,zhan2025towards,zhan2025epee}. In domains where data privacy is less restrictive, cloud-based infrastructure offers a practical solution for accessing the benefits of large-scale models. However, in the biomedical domain, patient confidentiality and data security are paramount~\cite{act1996health,price2019privacy}. Thus, cloud deployment is often infeasible~\cite{mehrtak2021security}. As a result, running LLMs locally remains the primary approach. This raises a critical challenge: how to efficiently deploy high-performing models in resource-constrained local environments without compromising performance?

\begin{figure}
    \centering
    \includegraphics[width=0.8\linewidth]{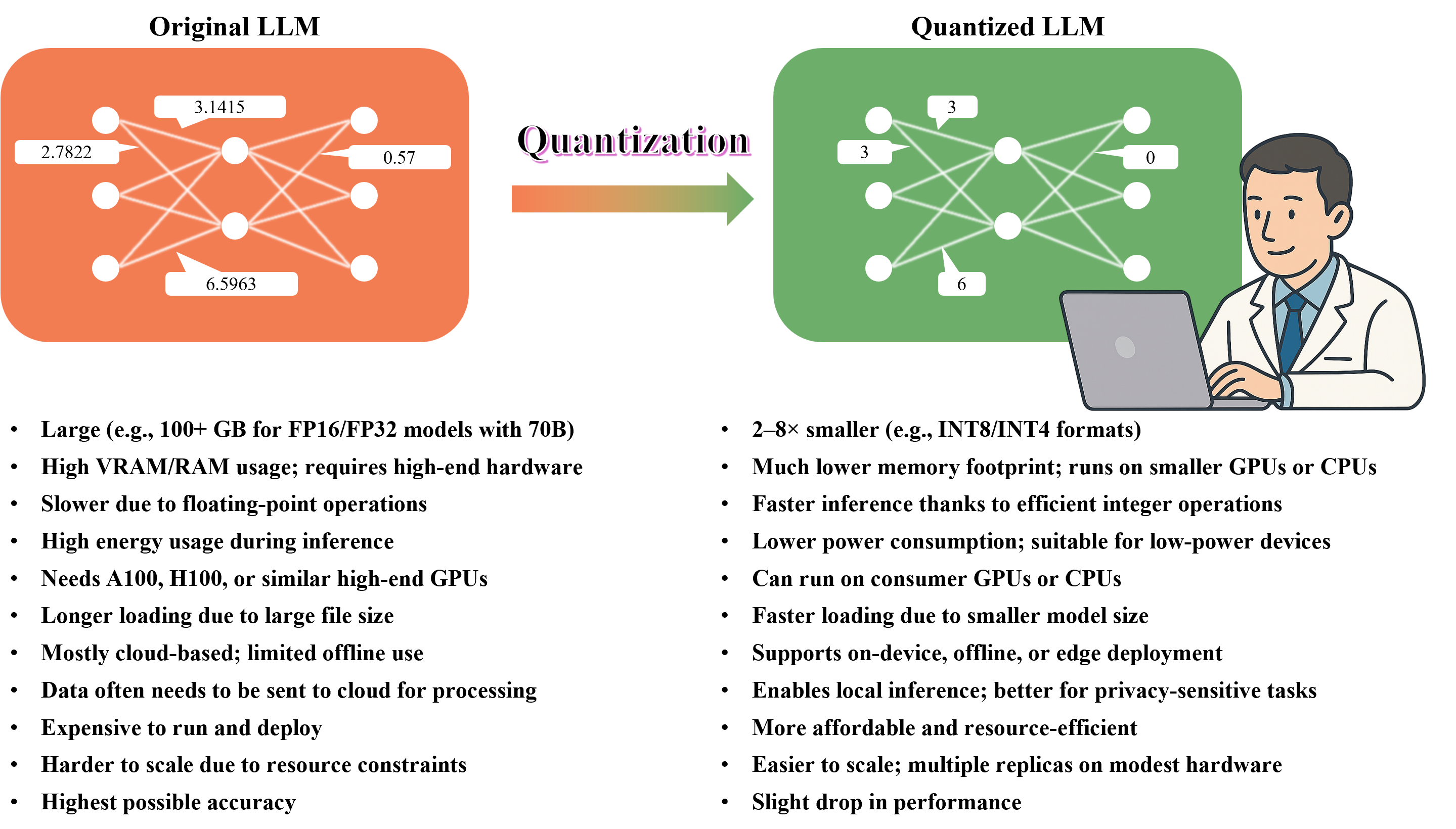}
    \caption{Comparison between default LLMs and their quantized counterparts.}
    \label{fig:overview}
\end{figure}

One promising approach to address this challenge is model quantization~\cite{gholami2022survey,deng2020model}, which reduces the precision of model weights, typically from 32- or 16-bit floating point representations to 8-bit or even lower formats, as shown in Figure \ref{fig:overview}. By compressing the model in this way, quantization significantly decreases the memory footprint and computational workload during inference, leading to faster execution times, lower power consumption, and the ability to run models on less powerful hardware such as consumer GPUs or edge devices~\cite{li2024evaluating,jin2024comprehensive}. These improvements are particularly valuable for deploying LLMs in biomedical settings, where data often cannot be transferred off-site due to strict privacy regulations, and where high-performance GPUs may not be readily available.
As such, quantization has emerged not only as a tool for efficiency but also as a critical enabler for the practical and responsible deployment of LLMs in real-world, resource-constrained environments~\cite{husom2025sustainable,girija2025optimizing}.

Despite the growing interest in model quantization, most existing studies have primarily focused on general-domain NLP tasks and benchmarks. For example, Liu et al.\cite{liu2024evaluating} and Li et al.\cite{li2024evaluating} conducted extensive evaluations of quantized LLMs on a variety of general-purpose NLP tasks, demonstrating their efficiency and competitive performance. Jin et al.\cite{jin2024comprehensive} provided a comprehensive analysis of instruction tuning for quantized LLMs and offered practical recommendations for deployment. Additionally, Li et al.\cite{li2025quantization} explored the reasoning capabilities of quantized LLMs in mathematical tasks, while Marchisio et al.\cite{marchisio2024does} investigated quantization strategies for multimodal models. Further studies have examined the combination of quantization with low-rank adaptation techniques such as LoRA for fine-tuning, as seen in works by Xu et al.\cite{xu2023qa} and Zhou et al.~\cite{zhou2025uncertainty}.
In contrast, the impact of quantized LLMs on biomedical tasks remains largely unexplored. Biomedical text presents unique challenges compared to general-domain corpora, including domain-specific terminology, more complex syntax, and highly specialized semantics. These differences mean that conclusions drawn from general-domain evaluations may not directly transfer to biomedical applications. Therefore, it is crucial to empirically assess the effectiveness, robustness, and trade-offs of quantized LLMs within the biomedical NLP landscape.

To address this gap in the literature, we propose a systematic evaluation of quantized large language models in the context of biomedical NLP tasks. Our goal is to assess the practical viability of using quantized LLMs in privacy-sensitive, resource-constrained biomedical environments and to provide actionable insights for researchers and practitioners seeking to adopt such models in real-world applications.
Specifically, we conduct comprehensive experiments across a variety of biomedical tasks, evaluating the performance of quantized models under different settings. These include varying degrees of quantization (e.g., 8-bit, 4-bit), different model architectures and sizes, and multiple common task types such as named entity recognition, relation extraction, question answering, and multi-label classification. We also investigate how quantized LLMs perform under few-shot learning conditions and in combination with different prompting strategies. By exploring these dimensions, we aim to establish a clearer understanding of the trade-offs between efficiency and task performance, and to offer practical guidance for selecting and deploying quantized LLMs in biomedical scenarios.

\section{Results}
In this section, we present our main results of analysis on quantized LLMs with a focus on several practical facets, including various tasks, scalability, model architectures and sizes, and advanced methods such as few-shot learning and prompting methods.

\subsection{Quantized LLMs on MLC, NER, RE, QA tasks}
\begin{figure}[htbp]
    \centering
    \includegraphics[width=0.99\linewidth]{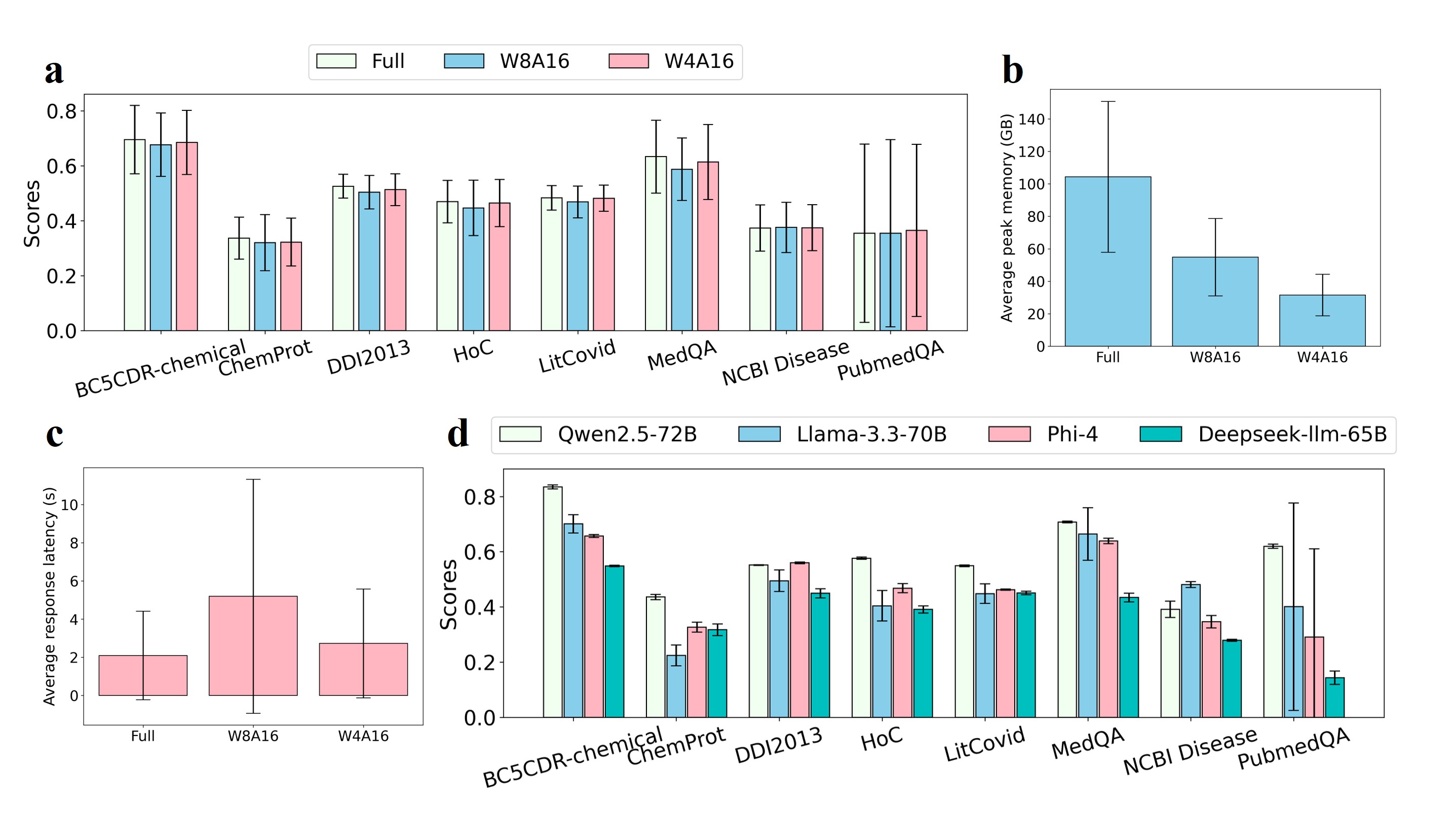}
    \caption{Quantitative comparison of different quantization schemes and LLMs. Bars indicate mean, error bars indicate standard deviation.}
    \label{fig: main}
\end{figure}

Table \ref{main_table} summarizes the performance of four LLMs across eight datasets covering four distinct tasks, under three levels of numerical precision.
Figure \ref{fig: main} presents a comparison of model performance, peak memory usage, and average response latency under different quantization settings. It also includes a performance comparison across different LLMs.
The results show that across four tasks, quantized models experience negligible performance degradation while achieving significant reductions in peak memory usage. However, quantization leads to an increase in response latency. Among the four models evaluated, the Qwen model demonstrates the strongest overall performance.

\begin{table}[htbp]
  \centering
  \caption{Performance and efficiency comparison of full-precision and quantized LLMs (W8A16: 8-bit weights with 16-bit activations; W4A16: 4-bit weights with 16-bit activations) across biomedical NLP tasks, including multi-label classification, named entity recognition, relation extraction, and question answering. Reported metrics include memory usage (Mem, in GB), inference latency (Late, in seconds), and task-specific scores (F1 / Accuracy).}
  \label{main_table}

\begin{tabular}{lcccc|ccc|ccc}
\toprule
\hline
Dataset - Task / Model & Metric & \multicolumn{3}{c|}{Full precision} & \multicolumn{3}{c|}{W8A16} & \multicolumn{3}{c}{W4A16} \\
 & & Mem & Late & Score & Mem & Late & Score & Mem & Late & Score \\
\hline
\multicolumn{5}{l|}{ \textbf{ HoC - Multi-label document classification } } & \multicolumn{3}{c|}{}\\

\quad Deepseek-llm-65B & Macro F1 & 126.5 & 4.57 & 0.391 & 65.61 & 10.4 & 0.404 & 36.87 & 6.84 & 0.377 \\
\quad Llama-3.3-70B & Macro F1 & 132.37 & 3.74 & 0.45 & 68.85 & 11.4 & 0.343 & 39.38 & 4.87 & 0.419 \\
\quad Phi-4 & Macro F1 & 28.41 & 1.05 & 0.459 & 16.08 & 2.27 & 0.458 & 10.72 & 1.45 & 0.487 \\
\quad Qwen2.5-72B & Macro F1 & 136.41 & 2.75 & 0.576 & 72.58 & 6.83 & 0.581 & 41.0 & 3.48 & 0.573 \\

\multicolumn{5}{l|}{ \textbf{ LitCovid - Multi-label document classification } } & \multicolumn{3}{c|}{}\\

\quad Deepseek-llm-65B & Macro F1 & 127.15 & 1.96 & 0.444 & 66.37 & 4.73 & 0.452 & 37.41 & 2.36 & 0.456 \\
\quad Llama-3.3-70B & Macro F1 & 133.08 & 1.47 & 0.481 & 69.7 & 3.55 & 0.41 & 39.95 & 2.0 & 0.452 \\
\quad Phi-4 & Macro F1 & 29.2 & 0.47 & 0.461 & 16.91 & 1.02 & 0.462 & 11.36 & 0.67 & 0.465 \\
\quad Qwen2.5-72B & Macro F1 & 137.13 & 1.27 & 0.546 & 73.46 & 2.97 & 0.548 & 41.6 & 1.67 & 0.553 \\

\multicolumn{5}{l|}{ \textbf{ BC5CDR-chemical - Named entity recognition } } & \multicolumn{3}{c|}{}\\

\quad Deepseek-llm-65B & Entity F1 & 125.97 & 6.48 & 0.547 & 64.92 & 16.58 & 0.547 & 36.41 & 8.95 & 0.552 \\
\quad Llama-3.3-70B & Entity F1 & 131.87 & 6.52 & 0.734 & 68.21 & 15.77 & 0.668 & 38.94 & 7.15 & 0.7 \\
\quad Phi-4 & Entity F1 & 27.78 & 1.97 & 0.656 & 15.23 & 4.66 & 0.663 & 10.23 & 2.82 & 0.653 \\
\quad Qwen2.5-72B & Entity F1 & 135.87 & 6.8 & 0.842 & 71.62 & 17.79 & 0.828 & 40.56 & 7.73 & 0.833 \\

\multicolumn{5}{l|}{ \textbf{  NCBI Disease - Named entity recognition } } & \multicolumn{3}{c|}{}\\

\quad Deepseek-llm-65B & Entity F1 & 125.89 & 6.22 & 0.279 & 64.82 & 15.95 & 0.275 & 36.34 & 9.09 & 0.282 \\
\quad Llama-3.3-70B & Entity F1 & 131.77 & 6.52 & 0.469 & 68.12 & 17.31 & 0.488 & 38.88 & 7.56 & 0.486 \\
\quad Phi-4 & Entity F1 & 27.68 & 1.96 & 0.332 & 15.1 & 4.74 & 0.336 & 10.16 & 2.93 & 0.372 \\
\quad Qwen2.5-72B & Entity F1 & 135.78 & 6.94 & 0.413 & 71.49 & 18.76 & 0.402 & 40.49 & 7.86 & 0.357 \\

\multicolumn{5}{l|}{ \textbf{ ChemProt - Relation extraction } } & \multicolumn{3}{c|}{}\\

\quad Deepseek-llm-65B & Macro F1 & 126.43 & 1.01 & 0.324 & 65.54 & 2.52 & 0.334 & 36.81 & 1.52 & 0.294 \\
\quad Llama-3.3-70B & Macro F1 & 132.3 & 0.82 & 0.249 & 68.73 & 1.64 & 0.181 & 39.3 & 1.09 & 0.243 \\
\quad Phi-4 & Macro F1 & 28.27 & 0.37 & 0.336 & 15.82 & 0.62 & 0.338 & 10.62 & 0.37 & 0.305 \\
\quad Qwen2.5-72B & Macro F1 & 136.33 & 0.74 & 0.435 & 72.53 & 1.67 & 0.427 & 40.95 & 1.03 & 0.446 \\

\multicolumn{5}{l|}{ \textbf{ DDI2013 - Relation extraction } } & \multicolumn{3}{c|}{}\\

\quad Deepseek-llm-65B & Macro F1 & 126.19 & 0.87 & 0.464 & 65.29 & 2.03 & 0.452 & 36.65 & 1.3 & 0.432 \\
\quad Llama-3.3-70B & Macro F1 & 132.03 & 0.83 & 0.524 & 68.48 & 1.51 & 0.45 & 39.14 & 1.12 & 0.51 \\
\quad Phi-4 & Macro F1 & 28.03 & 0.25 & 0.56 & 15.59 & 0.5 & 0.562 & 10.43 & 0.38 & 0.557 \\
\quad Qwen2.5-72B & Macro F1 & 136.09 & 0.82 & 0.552 & 72.24 & 1.86 & 0.55 & 40.78 & 1.18 & 0.553 \\

\multicolumn{5}{l|}{ \textbf{ MedQA -  Question answering } } & \multicolumn{3}{c|}{}\\

\quad Deepseek-llm-65B &  Accuracy & 126.75 & 0.75 & 0.441 & 65.89 & 1.77 & 0.445 & 37.08 & 1.07 & 0.416 \\
\quad Llama-3.3-70B &  Accuracy & 132.63 & 0.5 & 0.732 & 69.15 & 0.88 & 0.555 & 39.58 & 0.77 & 0.705 \\
\quad Phi-4 &  Accuracy & 28.67 & 0.25 & 0.648 & 16.29 & 0.53 & 0.64 & 10.94 & 0.39 & 0.628 \\
\quad Qwen2.5-72B &  Accuracy & 136.7 & 0.45 & 0.71 & 72.76 & 0.92 & 0.709 & 41.26 & 0.73 & 0.704 \\

\multicolumn{5}{l|}{ \textbf{ PubMedQA -  Question answering } } & \multicolumn{3}{c|}{}\\

\quad Deepseek-llm-65B & Accuracy & 126.44 & 0.7 & 0.166 & 65.5 & 1.58 & 0.146 & 36.82 & 1.1 & 0.118 \\
\quad Llama-3.3-70B & Accuracy & 132.34 & 0.63 & 0.706 & 68.78 & 1.07 & 0.758 & 39.33 & 0.89 & 0.756 \\
\quad Phi-4 & Accuracy & 28.31 & 0.2 & 0.61 & 15.87 & 0.4 & 0.532 & 10.66 & 0.34 & 0.602 \\
\quad Qwen2.5-72B & Accuracy & 136.37 & 0.48 & 0.616 & 72.46 & 0.92 & 0.614 & 40.97 & 0.8 & 0.628 \\
\hline
\bottomrule
\end{tabular}

\end{table}

\subsection{Domain quantized LLMs on MLC, NER, RE, QA tasks}
\begin{figure}[htbp]
    \centering
    \includegraphics[width=1\linewidth]{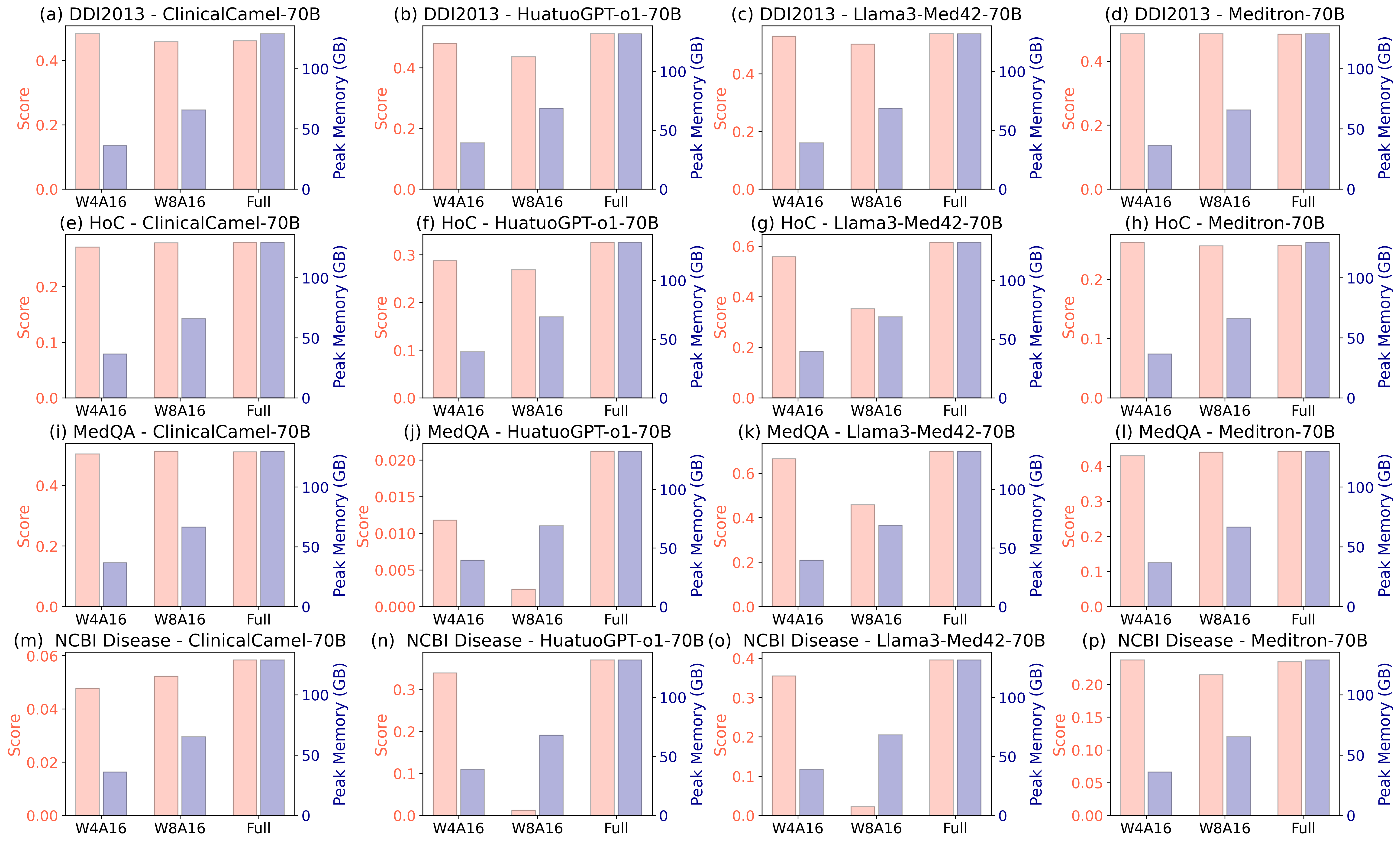}
    \caption{Comparison of different quantized LLMs from the biomedical domain on DDI2013, HoC, MedQA, and NCBI Disease datasets across MLC, NER, RE, and QA tasks. 
    }
    \label{fig: medicalLLM}
\end{figure}

An increasing number of models have been further trained with domain-specific knowledge, and many large models now inherently contain rich domain expertise. Therefore, it is essential to explore the effectiveness of quantization techniques on these further-trained models.
Figure \ref{fig: medicalLLM} compares the performance and memory consumption of the ClinicalCamel-70B, HuatuoGPT-o1-70B, Llama3-Med42-70B, and Meditron-70B models on the DDI2013, HoC, MedQA, and NCBI Disease datasets. As shown in the figure, quantized LLMs exhibit a significant reduction in GPU memory requirements across all cases. Notably, for the DDI2013 and HoC datasets, quantization has minimal impact on model performance, especially when quantized to 4 bits. However, in some cases, HuatuoGPT-o1-70B and Llama3-Med42-70B show substantial performance degradation when quantized to 8 bits.

\subsection{Quantized LLMs with different size}
\begin{figure}[htbp]
    \centering
    \includegraphics[width=1\linewidth]{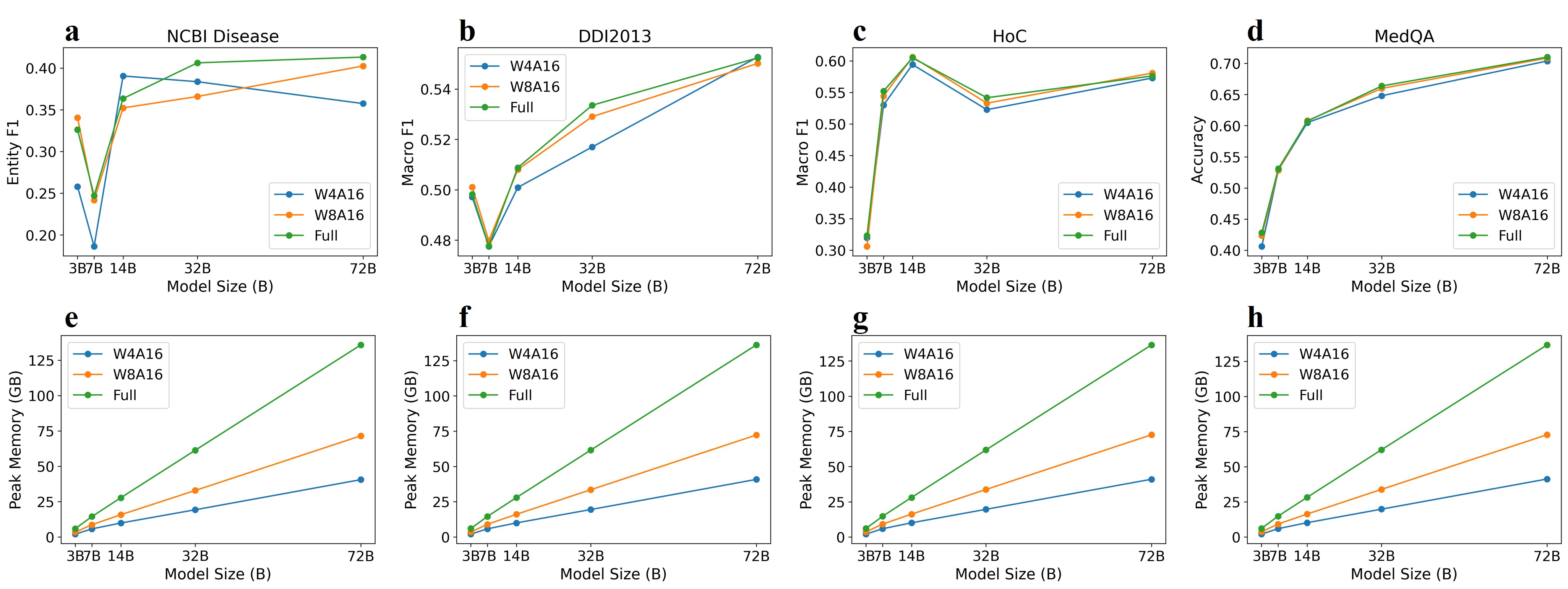}
    \caption{Comparison of various quantized Qwen models on DDI2013, HoC, MedQA, and NCBI Disease datasets across MLC, NER, RE, and QA tasks. 
    }
    \label{fig: size}
\end{figure}

The size of the model is a key factor to consider when deploying locally, as it directly impacts GPU memory requirements. Therefore, it is important to investigate how quantization techniques affect models of different sizes.
Figure \ref{fig: size} presents the performance (a–d) and memory usage (e–h) of Qwen2.5 models of various sizes under different quantization settings across four datasets: DDI2013, HoC, MedQA, and NCBI Disease. As the model size increases, memory usage grows approximately linearly. For the NCBI and HoC datasets, the performance of the 14B model has essentially plateaued, indicating saturation. In contrast, for the DDI and MedQA datasets, performance continues to improve with increasing model size. When comparing different quantization levels, the performance curves largely overlap, suggesting that quantized models can achieve comparable results to their full-precision counterparts. This highlights the effectiveness of quantization techniques in preserving model performance while reducing memory requirements.

\subsection{Quantized LLMs with few-shot learning}
\begin{figure}[htbp]
    \centering
    \includegraphics[width=0.6\linewidth]{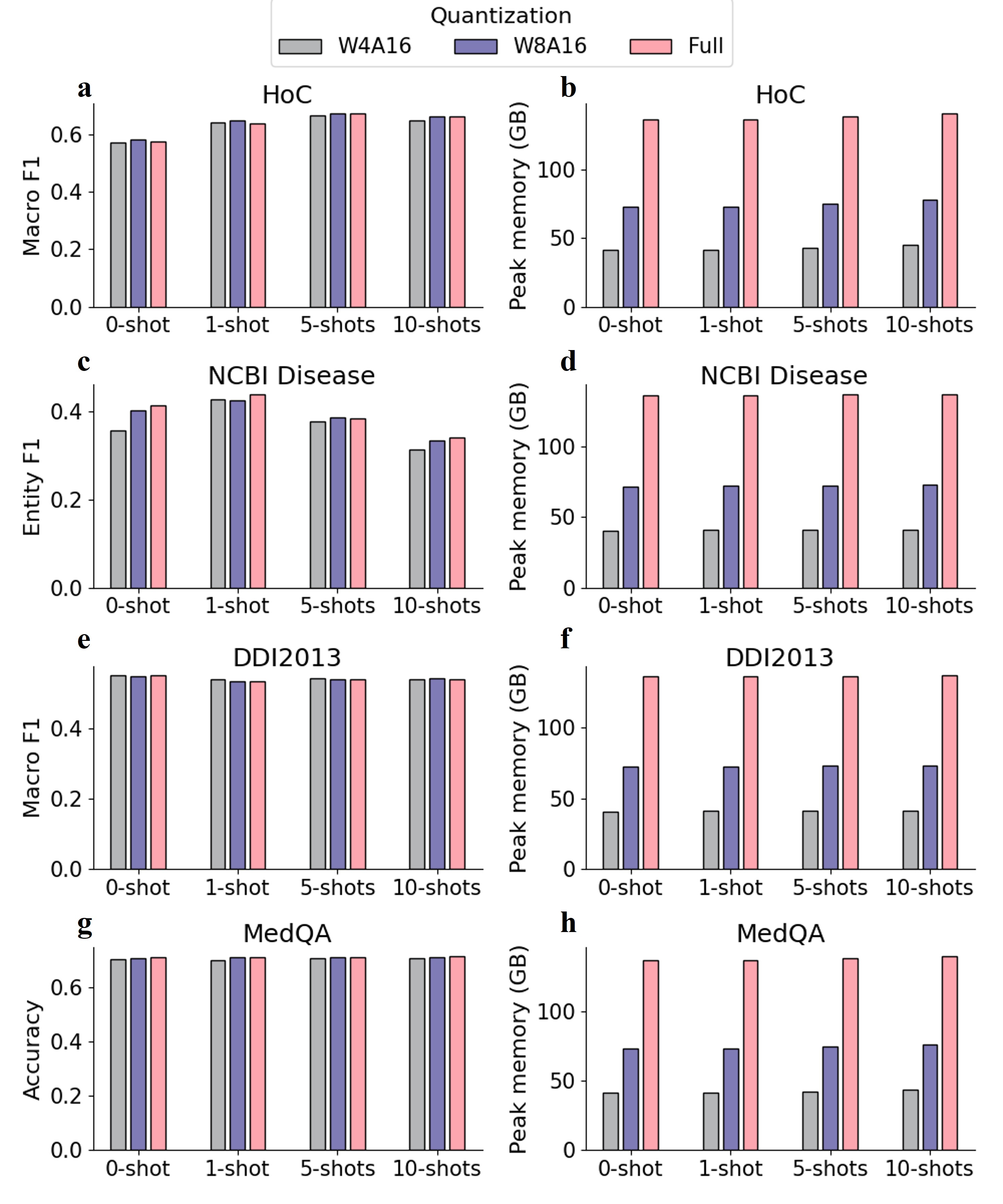}
    \caption{Comparison of quantized Qwen2.5-70B model with few-shot setting on DDI2013, HoC, MedQA, and NCBI Disease datasets across MLC, NER, RE, and QA tasks. 
    }
    \label{fig: fewshot}
\end{figure}

Few-shot learning continues to play an important role in the era of large language models (LLMs). It enables models that haven't been explicitly trained on a specific task to achieve relatively strong performance with just a few examples. However, few-shot learning also typically leads to longer prompts, which increases the demand for computational resources and enlarges the KV cache, placing a heavier burden on GPU memory. To better understand this trade-off, we conducted a series of investigations.
As shown in Figure \ref{fig: fewshot}, GPU memory usage increases with prompt length under different quantization settings.

\subsection{Prompting strategies on quantized LLMs}
\begin{figure}[htbp]
    \centering
    \includegraphics[width=0.8\linewidth]{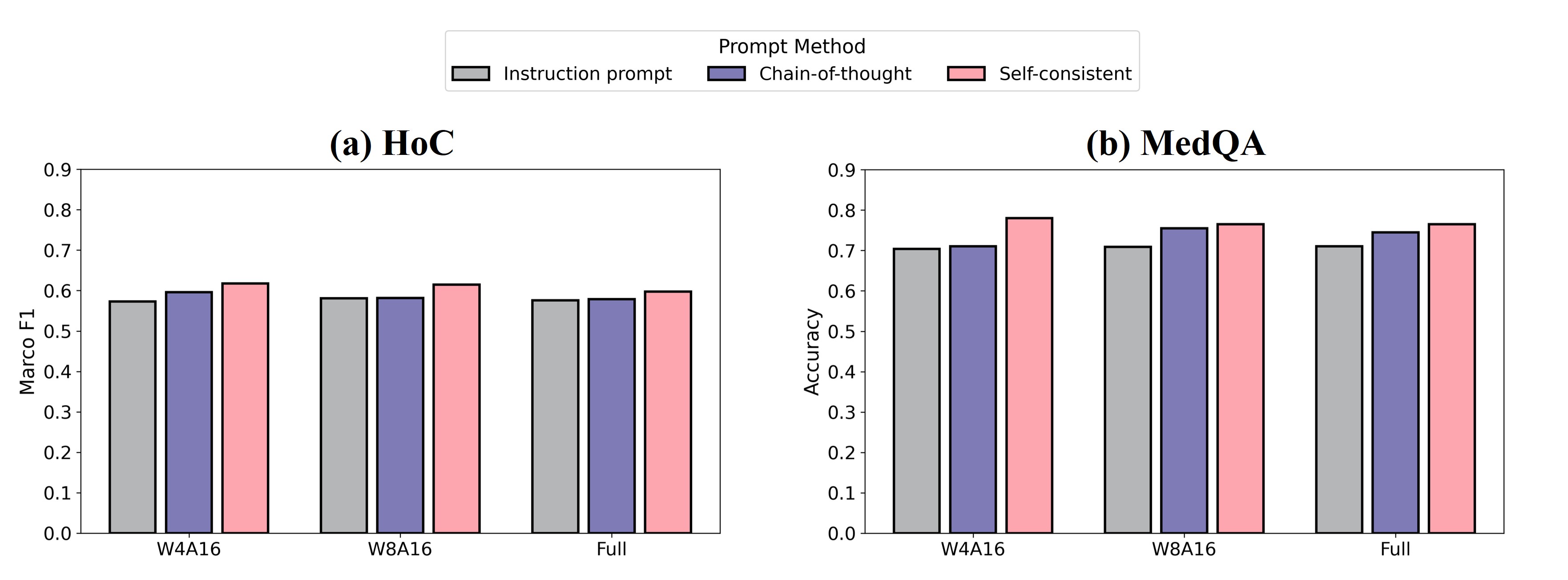}
    \caption{Performance comparison of the quantized Qwen2.5-70B model using instruction prompts, chain-of-thought prompts, and self-consistent prompting methods on the HoC and MedQA datasets.}
    \label{fig: prompt}
\end{figure}

Effective prompting of large language models (LLMs) has emerged as a critical and increasingly prominent research area. As LLMs continue to scale, making them more resource-intensive to train, prompt engineering offers a cost-efficient means of leveraging their capabilities. Accordingly, this work investigates the impact of quantization on LLM performance under consistent prompts or prompting strategies.
Figure \ref{fig: prompt} illustrates the performance of three prompting approaches—instruction prompting, chain-of-thought (CoT) prompting, and self-consistent prompting. The results demonstrate that more sophisticated prompting methods yield substantial performance improvements. Notably, these trends are consistently observed across different quantization levels, indicating that quantized LLMs remain compatible with and responsive to advanced prompting techniques.

\section{Discussion}

In this study, we conducted a systematic evaluation of quantization techniques applied to large language models (LLMs). As LLMs continue to scale rapidly, quantization has become an essential strategy for enabling their deployment in real-world scenarios, especially for small companies, research labs, clinics, and edge devices that cannot keep up with the increasing computational demands. This is particularly crucial in biomedicine, where patient privacy is non-negotiable, and running high-performance models locally—on secure, often resource-constrained devices—remains a persistent challenge. Our paper benchmarks 12 LLMs, including both general-purpose and biomedical-specific models, and various model sizes, across 8 datasets and 4 core tasks, while also exploring different prompting methods and few-shot learning setups. We believe this evaluation provides meaningful guidance for the broader research community working at the intersection of LLM deployment and biomedical applications.

Our results show that quantized LLMs maintain strong performance across all evaluated tasks—namely named entity recognition (NER), relation extraction (RE), multi-label classification (MLC), and question answering (QA). A major benefit of quantization is the substantial reduction in GPU memory requirements. Quantizing models to 8-bit typically reduces memory usage by half, and 4-bit quantization brings it down to nearly a quarter of the original footprint. With such improvements, it becomes feasible to run a 70-billion-parameter model on a 40GB GPU, making high-capacity models accessible even on consumer-grade hardware. More importantly, this reduction in memory does not come at the cost of performance—quantized models perform comparably to their full-precision counterparts. However, we do observe a moderate increase in latency. This is likely due to our design choice of using full-precision computation during inference to preserve model accuracy, which introduces overhead in the form of precision conversion. Nevertheless, quantization proves highly effective for biomedical use, enabling large-scale models to be deployed in resource-limited settings with minimal compromise on capability.

We also found that quantization has little to no detrimental impact on domain-specific models. For example, models such as ClinicalCamel, which are trained extensively on biomedical data, retain their specialized knowledge after quantization. This suggests that the medical understanding embedded in these models is preserved and not lost during the quantization process. Contrary to concerns that such models may suffer reduced generalization or forget domain-specific information, our findings show they remain robust and reliable. This reinforces the case for broader application of quantization in biomedical AI systems, especially where domain relevance and trustworthiness are crucial.

Furthermore, our experiments demonstrate that quantization remains compatible with other modern techniques commonly used in LLM deployment, such as few-shot learning and prompt engineering. When combined with these approaches, quantized models continue to exhibit strong generalization, instruction-following, and reasoning capabilities. Despite a slight drop in performance in some cases, the trade-off is minimal and acceptable, especially given the drastic improvements in memory efficiency. This makes it possible to deliver nearly the same user experience with quantized models as with full-precision ones, even in settings with limited computing power. For the biomedical field, this is particularly valuable—it shows that powerful, instruction-tuned LLMs can be safely and effectively used in practice without requiring high-end infrastructure.

Despite these findings, our study has several limitations. Due to limited computing resources, our evaluation focused primarily on models with approximately 70 billion parameters, and we were unable to include even larger models exceeding 100 billion parameters. Additionally, our latency measurements were based on actual runtime, which means the results may have been influenced by server load or background processes during testing. As such, these latency figures should be interpreted as approximate and used mainly for reference.

\begin{figure}
    \centering
    \includegraphics[width=0.99\linewidth]{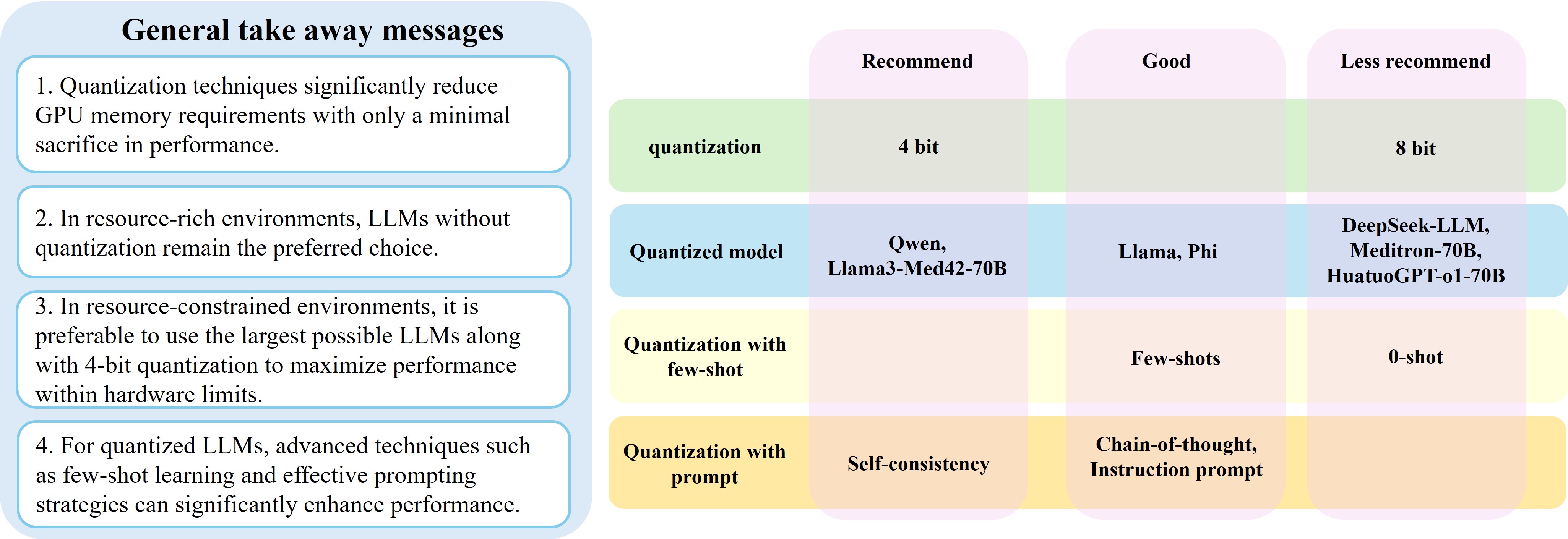}
    \caption{Recommendations for using quantized LLMs in biomedical natural language processing applications.}
    \label{fig:recommendation}
\end{figure}

In conclusion, our evaluation helps to fill an important research gap by assessing the practical value of quantization in biomedical contexts. We hope that our findings—and the usage recommendations summarized in Figure \ref{fig:recommendation}—will serve as a useful guide for clinicians and biomedical researchers seeking to adopt LLMs in real-world, resource-constrained environments. Quantization not only enables broader accessibility to powerful models but also helps ensure that privacy, efficiency, and performance can be maintained in sensitive applications like healthcare.

\section{Methods}

\subsection{Datasets}

To enhance the generalizability and practical value of our research, we selected four commonly studied tasks: named entity recognition (NER), relation extraction (RE), multi-label classification (MLC), and question answering (QA). For each task, we further chose two representative publicly available datasets: HoC~\cite{baker2016automatic} and LitCovid~\cite{chen2022multi} for MLC, BC5CDR~\cite{li2016biocreative} and NCBI~\cite{dougan2014ncbi} for NER, ChemProt~\cite{krallinger2017overview} and DDI~\cite{segura2013semeval} for RE, and MedQA~\cite{jin2021disease} and PubMedQA~\cite{jin2019pubmedqa} for QA. Depending on the characteristics of each task, we applied task-specific evaluation metrics. All datasets are summarized in Table \ref{tab:datasets}.
\begin{table}[ht]
\centering
\caption{Overview of the evaluation datasets and evaluation metrics.}
\begin{tabular}{llrl}
\hline
\textbf{Task} & \textbf{Dataset} & \textbf{size} & \textbf{Metric} \\
\hline
\multirow{2}{*}{Multi-label Classification} 
  & HoC               & 315   & Macro F1 \\
  & LitCovid          & 2,500  & Macro F1 \\
\hline
\multirow{2}{*}{Named Entity Recognition} 
  & BC5CDR-chemical  & 4,797 & Entity-level F1 \\
  & NCBI-disease      & 940  & Entity-level F1 \\
\hline
\multirow{2}{*}{Relation Extraction} 
  & ChemProt         & 16,943 & Macro F1 \\
  & DDI2013          & 5,761  & Macro F1 \\
\hline
\multirow{2}{*}{Question Answering} 
  & MedQA 5-option    & 1,273  & Accuracy \\
  & PubMedQA         & 500   & Accuracy \\
\hline
\end{tabular}
\label{tab:datasets}
\end{table}


\subsection{LLMs}

We evaluate a selection of representative large language models (LLMs) from both general-purpose and biomedical domains. Specifically, the general-purpose models include Llama-3.3-70B~\cite{dubey2024llama}, Qwen2.5-72B~\cite{qwen2025qwen25technicalreport}, Phi-4~\cite{abdin2024phi}, and Deepseek-LLM-65B~\cite{bi2024deepseek}. For domain-specific models, we include ClinicalCamel-70B~\cite{toma2023clinical}, Llama3-Med42-70B~\cite{christophe2024med42}, HuatuoGPT-o1-70B~\cite{chen2024huatuogpt}, and Meditron-70B~\cite{chen2023meditron}. To assess performance across different model scales, we also benchmark a range of Qwen2.5~\cite{qwen2025qwen25technicalreport} models with varying sizes: Qwen2.5-3B, Qwen2.5-7B, Qwen2.5-14B, Qwen2.5-32B, and Qwen2.5-72B.

\subsection{Quantization Setup}

To evaluate the impact of quantization on model performance, we apply a standard quantization setup using the Hugging Face Transformers library~\cite{wolf2019huggingface}. Specifically, we support both 4-bit and 8-bit quantization through the BitsAndBytes module~\cite{dettmers20218,dettmers2022llm}. For 4-bit quantization, we adopt the NF4 quantization scheme with float16 as the computation data type and enable double quantization to improve numerical stability. For 8-bit quantization, we use the default 8-bit loading configuration.

\subsection{Prompt Methods}

To investigate the interaction between quantization techniques and prompting methods, we primarily adopt instruction prompts~\cite{brown2020language,wei2021finetuned} in our experiments. These prompts explicitly describe the task, expected output format, and provide the input to guide the model. Additionally, to explore the impact of different prompting strategies, we examine whether quantized models can still follow reasoning instructions and benefit from performance improvements when using approaches such as chain-of-thought~\cite{wei2022chain} and self-consistency prompting~\cite{wang2022self}.

\subsection{Experimental Settings}

Our code is implemented based on the Hugging Face \textit{transformers} library~\cite{wolf2019huggingface}. All experiments are conducted primarily on four NVIDIA A100 GPUs, each with 40 GB of memory.
We use the BitsAndBytes library~\cite{dettmers20218,dettmers2022llm} to quantize LLMs, converting them from full precision to 8-bit or 4-bit integer formats. Despite the quantization, computations are internally performed in full precision, which helps maintain model accuracy during the forward pass.
To optimize computational resources and reduce runtime, we dynamically adjust the output token length depending on the task. For instance, classification tasks typically restrict the output to around 50 tokens, whereas named entity recognition tasks may require generating complete sentences, allowing up to 200 tokens.
All other model parameters follow the default configurations provided by the Hugging Face library.

\section{Data Availability}
The datasets used in this paper follow a previous benchmark study~\cite{chen2025benchmarking} and are available at \href{https://doi.org/10.5281/zenodo.14025500}{https://doi.org/10.5281/zenodo.14025500}.

\section{Code Availability}
The code is publicly available at \href{https://github.com/Learner4everrr/Benchmark_quantization}{https://github.com/Learner4everrr/Benchmark\_quantization}.

\bibliography{0_main}

\end{document}